\documentclass[letterpaper, 10 pt, conference]{ieeeconf}  

\IEEEoverridecommandlockouts                              

\usepackage{color}
\usepackage{graphicx} 
\usepackage[american]{babel}
\usepackage{subfigure}		
\usepackage{balance}
\usepackage{cite}
\usepackage{xcolor}
\usepackage{hyperref}

\title{\LARGE \bf
CinemAirSim: A Camera-Realistic Robotics Simulator for Cinematographic Purposes
}

\author{\centering Pablo Pueyo, Eric Cristofalo, Eduardo Montijano, and Mac Schwager
\thanks{This work was supported by a DGA scholarship and a NDSEG fellowship;; Spanish projects PGC2018-098817-A-I00 and PGC2018-098719-B-I00 (MCIU/AEI/FEDER, UE), DGA T04-FSE; NSF grants CNS-1330008 and IIS-1646921; ONR grant N00014-18-1-2830 and ONRG-NICOP-grant N62909-19-1-2027; and the Ford-Stanford Alliance Program.}
\thanks{P. Pueyo and E. Montijano are associated with the Instituto de Investigaci\'on en Ingenier\'ia de Arag\'on, Universidad de Zaragoza, Spain 
\texttt{\small \{ppueyor, emonti\}@unizar.es}}
\thanks{E. Cristofalo and M. Schwager are associated with the Department of Aeronautics and Astronautics, Stanford University, USA
\hspace{10mm}
\texttt{\small \{ecristof, schwager\}@stanford.edu}}
}

\begin{document}

\maketitle
\thispagestyle{empty}
\pagestyle{empty}

\begin{abstract}

Unmanned Aerial Vehicles (UAVs) are becoming increasingly popular in the film and entertainment industries, in part because of their maneuverability and perspectives they enable. 
While there exists methods for controlling the position and orientation of the drones for visibility, other artistic elements of the filming process, such as focal blur, remain unexplored in the robotics community.
The lack of cinematographic robotics solutions is partly due to 
the cost associated with the cameras and devices used in the filming industry, but also because state-of-the-art photo-realistic robotics simulators only utilize a full in-focus pinhole camera model which does not incorporate these desired artistic attributes.
To overcome this, the main contribution of this work
is to endow the well-known drone simulator, AirSim, with a cinematic camera as well as extend its API to control all of its parameters in real time, including various filming lenses and common cinematographic properties. 
In this paper, we detail the implementation of our AirSim modification, \textit{CinemAirSim}, present examples that illustrate the potential of the new tool, and highlight the new research opportunities that the use of cinematic cameras can bring to research in robotics and control.
\center{\href{https://github.com/ppueyor/CinematicAirSim}{github.com/ppueyor/CinematicAirSim}}
\end{abstract}

\section{Introduction}
The use of Unmanned Aerial Vehicles (UAVs) is wide-spread in industry and robotics research \cite{valavanis2015handbook}.
Among the list of potential uses of UAVs, there is growing interest in the film industry for those seeking a more unique cinematic experience.
Drones enable the recording of scenes and perspectives that are typically inaccessible by typical filming cranes or other infrastructure.

\begin{figure}[!ht]
\centering
\begin{tabular}{cc}
    \includegraphics[width=0.44\columnwidth,height=2.6cm]{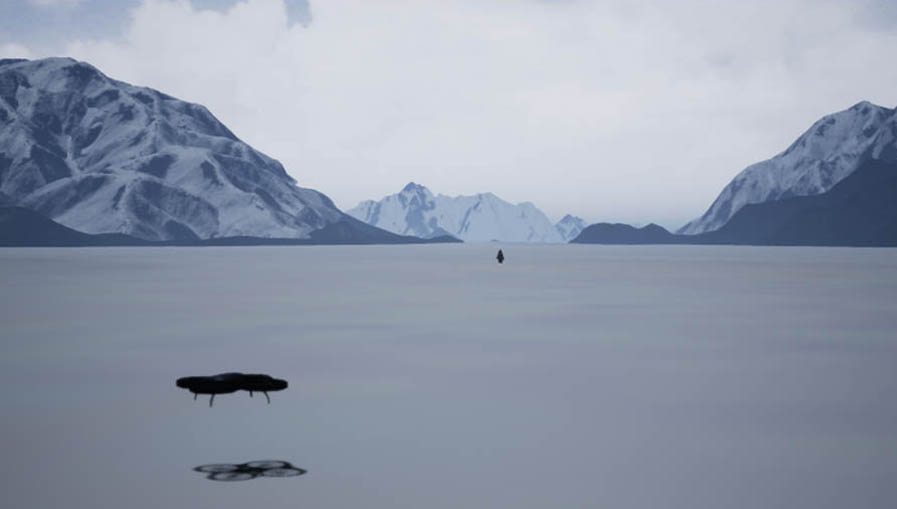} &
    \includegraphics[width=0.44\columnwidth,height=2.6cm]{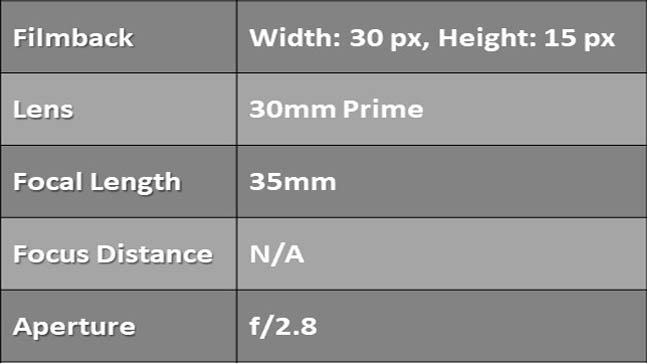}
    \\
    \includegraphics[width=0.44\columnwidth,height=2.6cm]{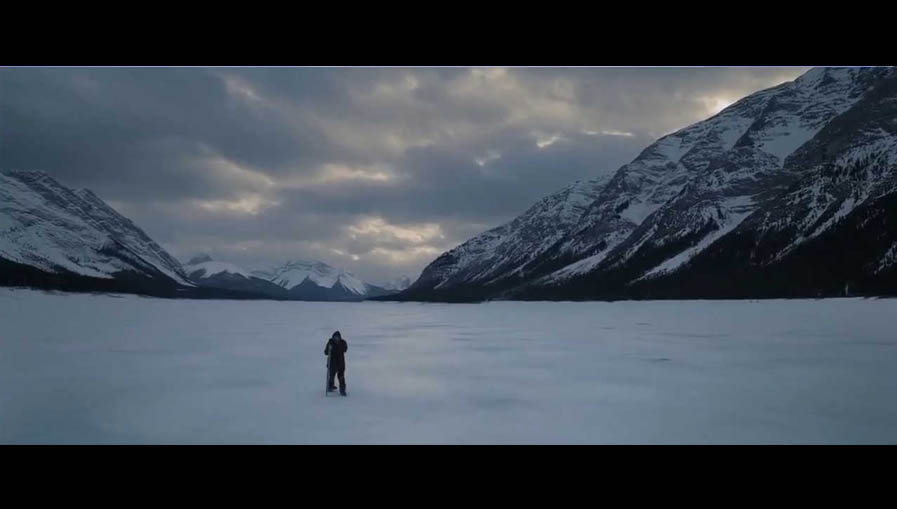} &
    \includegraphics[width=0.44\columnwidth,height=2.6cm]{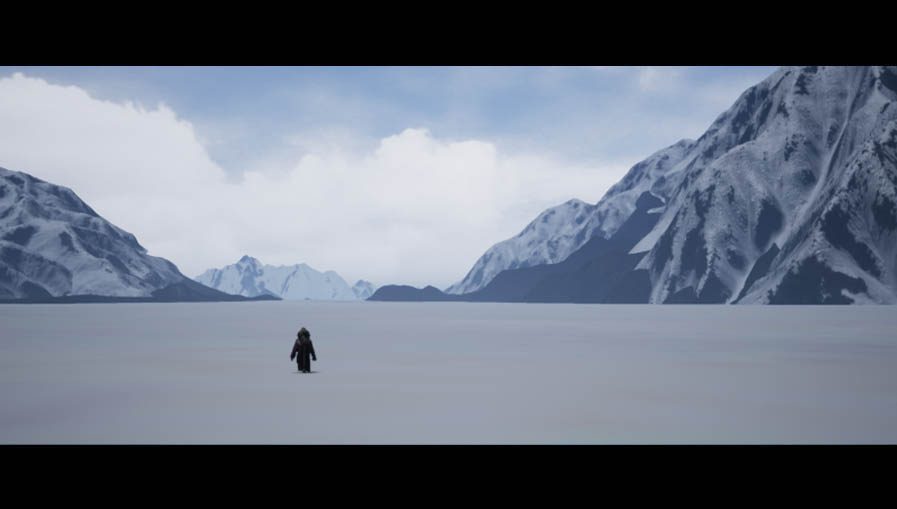}
\end{tabular}
\caption{\footnotesize CinemAirSim allows to simulate a cinematic camera inside AirSim. Users can control lens parameters like the focal length, focus distance or aperture, in real time. The figure shows a reproduction of a scene from ``The Revenant''. We illustrate the filming drone (top left), the current camera parameters (top right), the movie frame (bottom left) and simulated frame from the camera (bottom right). The supplementary video contains this scene and other experiments using the simulator.}
\label{fig:main}
\end{figure}

In the robotics community, there is interest in developing control and planning algorithms for cinematographic purposes.
For example, model predictive control is used in~\cite{nageli2017real} to calculate trajectories when filming a desired aerial shot. Different interfaces to generate paths according to artistic requests are presented in~\cite{gebhardt2016airways,huang2018act,huang2018through}. 
A high-level API that incorporates camera path planning and autonomous drone control for shooting purposes is described in~\cite{fleureau2016generic}. 
A $3$D tool that eases the shooting and the design of trajectories of UAVs for non-robotic experts 
is developed in~\cite{joubert2015interactive}. 
Visibility constraints, in terms of the focal length are discussed in~\cite{karakostas2020shot}.
Finally, autonomous key frame selection during the filming is discussed in~\cite{galvane2017automated}. 
All these works are focused on the design of trajectories, i.e., sequences of desired positions and orientations, of the drones for filming purposes.

Other works focus on the aesthetic and artistic issues associated to photographic aspects. For example,~\cite{bonatti2018autonomous} shows a system for aerial filming that is able to record scenes following some cinematographic principles, while~\cite{bonatti2020autonomous} uses deep reinforcement learning to present an autonomous drone platform that is able to follow an actor while getting a visually pleasant shot, based on aesthetic quality parameters.
However, to the best of our knowledge, there is no prior work that explores the control of the lens of the cameras to achieve better aesthetic and artistic outcomes associated to photographic aspects.
This is probably due to the additional complications associated with modeling cinematographic cameras, not to mention the expensive price of these sensors. 
Including cinematographic cameras in an existing robotics simulator could be a simple and cheap way to avoid these complications (see our solution to this problem in Fig.~\ref{fig:main}).

The research progress in robotics would be almost impossible without some software to virtually reproduce the robots as close to reality as possible. 
There are many examples of robotics simulators that have been used and improved for decades.
Stage and Gazebo~\cite{koenig2004design} are two of the most popular state-of-the-art simulators, thanks to their native integration with the Robotic Operating System (ROS). Webots~\cite{michel2004cyberbotics} is another interesting platform that includes a variety of robotic models and sensors and an advanced physics engine.
However, in order to account for the cinematographic aspects of filming, it is also very important that simulators have a powerful graphic engine capable of rendering photo-realistic scenes. 
Despite the features that make these simulators useful for general robotics --- their powerful physics engines, their easy connection with ROS, or the wide community of users that makes development easier --- the lack of realism in their graphic engines limits their potential in the context of cinematography.

The last few years have witnessed the appearance of new simulators that make use of state-of-the-art graphic engines, such as Unreal~\cite{Unreal} or Unity~\cite{Unity}. 
On one hand, FlightGoggles~\cite{guerra2019flightgoggles}
is a photo-realistic simulator able to render realistic environments thanks to the Unity Engine, as well as its capability of gathering data and the integration of lifelike simulated sensors. On the other hand, AirSim~\cite{shah2018airsim}, developed by Microsoft Research, is built upon Unreal Engine. 
Both simulators meet all the necessary requirements for the cinematographic purposes, but currently none of them offer the possibility of online and real-time modification of camera parameters associated with standard filming photographic effects, i.e., focal length, focus distance, and aperture.

The main contribution of this paper is a modification of AirSim, \textit{CinemAirSim}, that empowers the simulator with the artistic possibilities of a real filming camera. The existing AirSim API is extended with new methods that enable users to access and control a camera's focal length, focal distance, or the lens aperture. Our contribution allows different types of lenses to be selected and considered in the control problem by exploiting the potential of the Unreal Engine.
All the files required to modify the original simulator can be downloaded for free from the Github page of the project\footnote{\href{https://github.com/ppueyor/CinematicAirSim}{github.com/ppueyor/CinematicAirSim}}.
The paper also includes examples where the different parameters are adjusted during the drone's flight to mimic scenes from movies awarded with the Academy Award for Best Cinematography in different years.

The rest of the paper is organized as follows. Section II describes the two main software elements of the simulator, Airsim and Unreal. Section III gives all the details of the integration of the cinematic camera into AirSim. Section IV shows different examples using the simulator to control both the drone and the camera parameters in order to reproduce real movie scenes. Finally, Section V presents the conclusions of the work and discusses the future lines of research that our software can enable in the robotics community.

\section{AirSim and Unreal}

\begin{figure}[!b]
\centering
\includegraphics[width=1\linewidth]{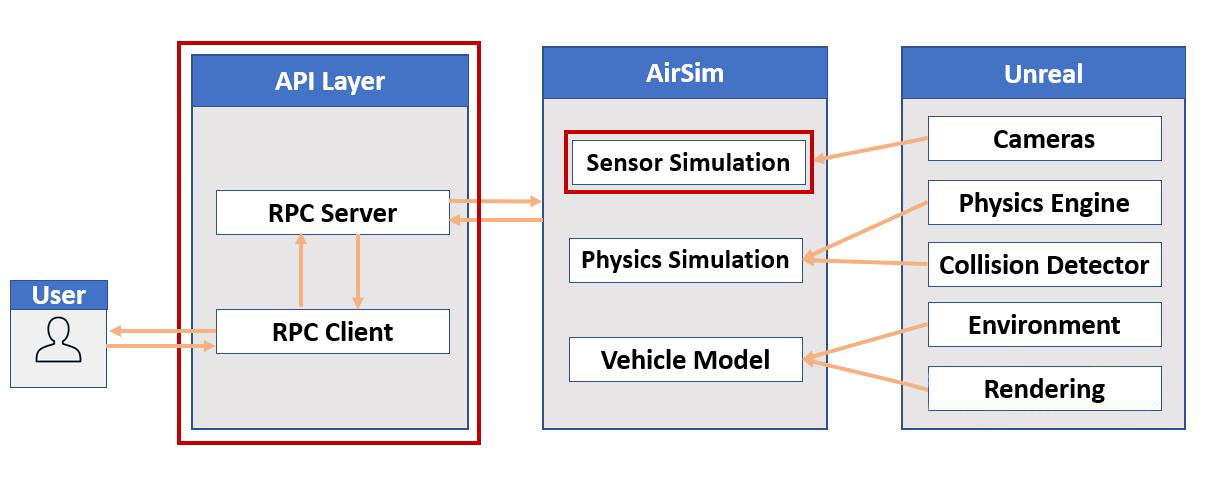}
\caption{\footnotesize Main software components of the programs used in CinemAirSim.}
	\label{fig:software_architecture}
\end{figure}

In this section we describe the two main programs that sustain CinemAirSim: the graphics engine Unreal and the robotics simulator AirSim.
The main software components of these programs are schematized in Fig.~\ref{fig:software_architecture}.

\subsection{Unreal}
The Unreal Engine is a graphics engine that was developed by Epic Games and is widely used in the video game industry. 
Its powerful graphic capabilities, open-source license, and the ability to create virtual environments make this tool extremely appealing for research.

Since the principal aim of the cinematographic simulator presented in this paper is to enable robotics research with the ability to acquire artistic shots, the quality of the visualization is an important aspect in its development.
In that sense, Unreal comes with a powerful graphics engine, which has a high number of characteristics that will be important to the aspirations of the simulator. 
In particular, it is worth-mentioning the powerful rendering capabilities, the visualization during the execution of the play using multiple configurable cameras, and the high fidelity simulation of the light, weather, and reflections of scene objects.

In addition, the environment editor of Unreal offers wide capabilities to create these realistic scenes. There are existing tools that allow users to design environments that are very close to reality in a quick and user-friendly way. 
There also exists an extensive community of active users that are creating $3$D models and maps available to download for free. 
All of this simplifies the task of creating custom scenarios to the robotics researchers.
Finally, Unreal comes with C++ code that allows to move the elements, direct the cameras, or control the light sources of the scene in real time.
Moreover, there are some works that provide an easy integration of Unreal with ROS~\cite{mania19scenarios} and OpenCV~\cite{qiu2016unrealcv}.

\subsection{AirSim}
AirSim (Aerial Informatics and Robotics Platform) was launched in 2017~\cite{shah2018airsim} by Microsoft Research as a plugin for the Unreal Engine 4. 
Like Unreal, AirSim also has an extensive community of users that are actively improving and discussing potentially interesting features of the simulator. Moreover, the possibility of generating binaries of code that integrates all the elements (environments, configurations) in a ready-to-run file enables AirSim to be used with a specific purpose by those users that are not so familiar to the robotics scope, such as cinematographers in the film industry. This is the method that Microsoft followed for the drone competition \textit{Game of Drones} during NeurIPS 2019~\cite{madaan2020airsim}.


AirSim provides a realistic simulation of a generic quadrotor and offers the possibility to customize it to accommodate to different commercial models.
The design of AirSim takes into account diverse factors in order to achieve the best realism when emulating the behaviour of actual drones. 
The model of the vehicles and its actuators, the effect of the environment variables such as gravity, magnetic field and air pressure on the robots, or the accelerations when moving, to name a few.
Moreover, the simulator uses the physics engine included in Unreal to detect and model collisions.

While several models of sensors are embedded to the robots, given the scope of the paper, we focus on the vision sensors and describe the current properties of these sensors in AirSim. 
The default camera provided by AirSim is called ``PIPCamera'' due to the possibility to simulate different vision sensors within the same layout, i.e., RGB image, depth perspective, or object segmentation.
AirSim allows to attach to the vehicle several of these cameras at different locations.
The visualization of the images in AirSim can be done in two ways: i) using the main viewport of Unreal and ii) using a \emph{SceneCaptureComponent}.
This component is also used to acquire the images of the cameras for further post processing.
More importantly for the cinematic purpose, the main intrinsic configurations are the resolution (the sensor's width and height) and the field-of-view, which can only be modified offline in the program settings.
This type of camera follows the classic pinhole model, which limits its use on the cinematographic scope when compared to more complex optic models, like the thin-lens models.


Finally, all these elements and parameters are controlled through the provided API. When the execution of the flight starts, a Remote Procedure Call (RPC) server is launched. This server allows the user to make RPC calls to control the different drones in real time. 



\section{Cinematographic Cameras in AirSim}

\begin{figure*}[!h]
\addtolength{\tabcolsep}{-3pt} 
    \begin{tabular}{ccccc}

         \includegraphics[width=0.19\linewidth,height=2.25cm]{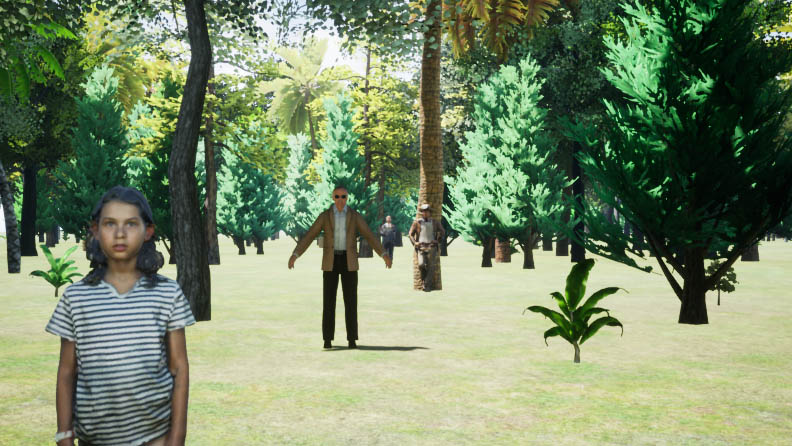}
         &
          \includegraphics[width=0.19\linewidth,height=2.25cm]{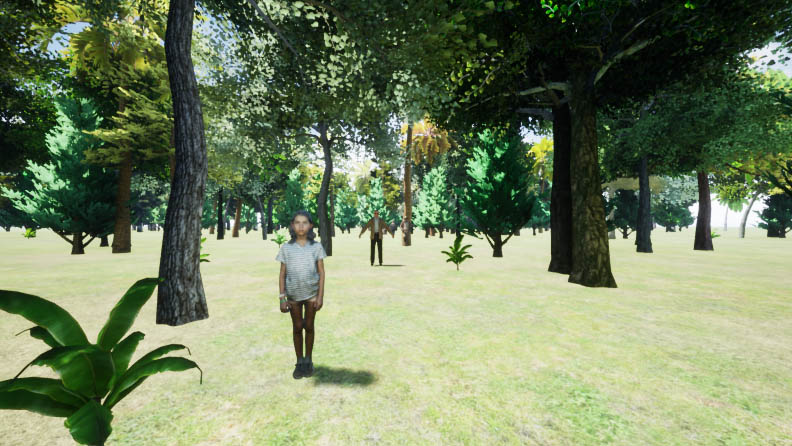}
         &
         \includegraphics[width=0.19\linewidth,height=2.25cm]{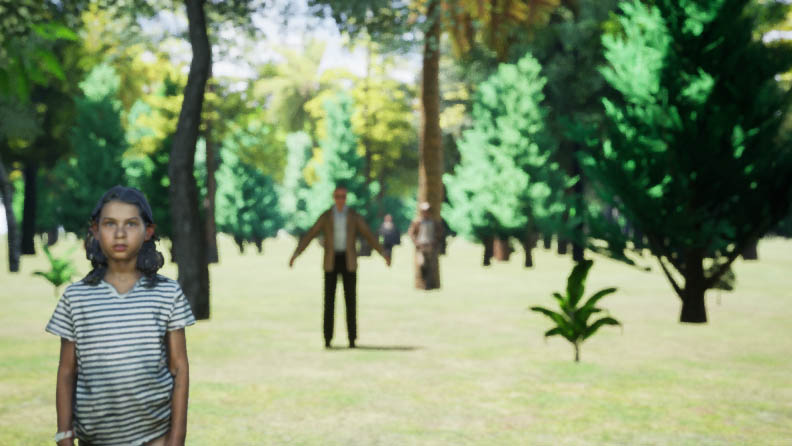}
         &
          \includegraphics[width=0.19\linewidth,height=2.25cm]{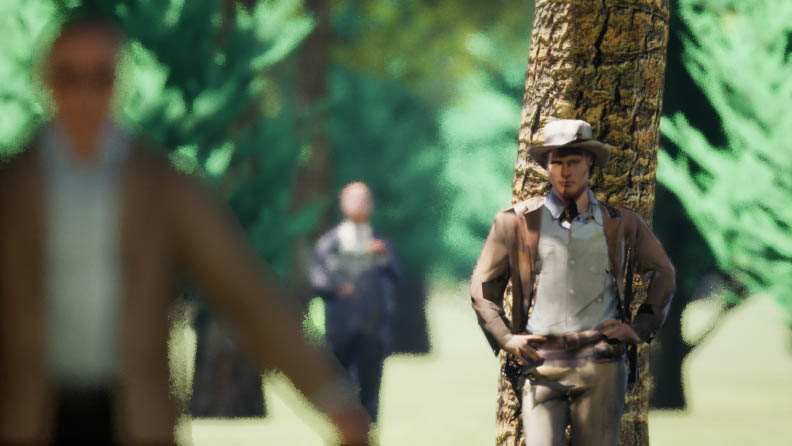}
         &
          \includegraphics[width=0.19\linewidth,height=2.25cm]{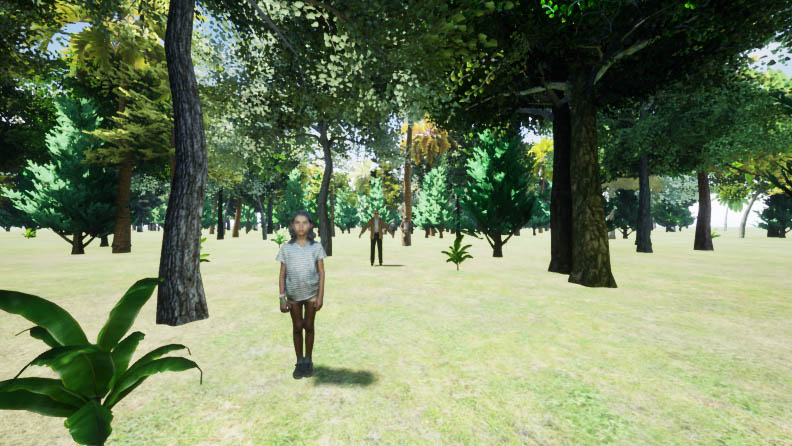}
         \\
         \small Filmback 16:9 DSLR &
         \small Lens: 12mm Prime f/2.8 &
         \small Focus distance: 300cm &
         \small Aperture: f/1.2 &
         \small Focal Length: 12mm\\[5pt]
         \includegraphics[width=0.19\linewidth,height=2.25cm]{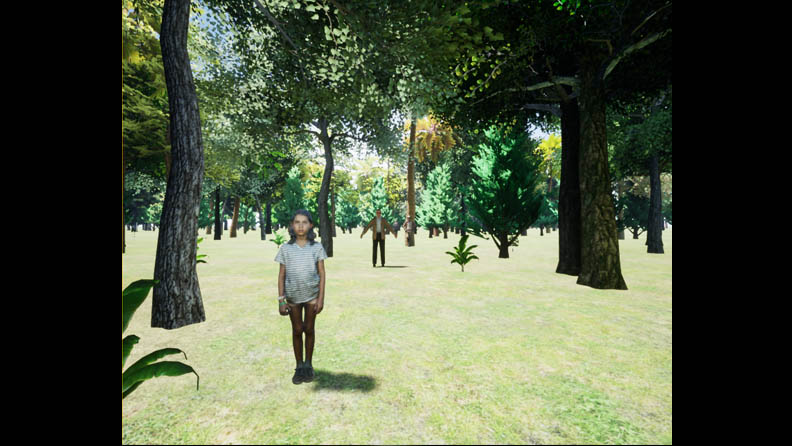}
         &
          \includegraphics[width=0.19\linewidth,height=2.25cm]{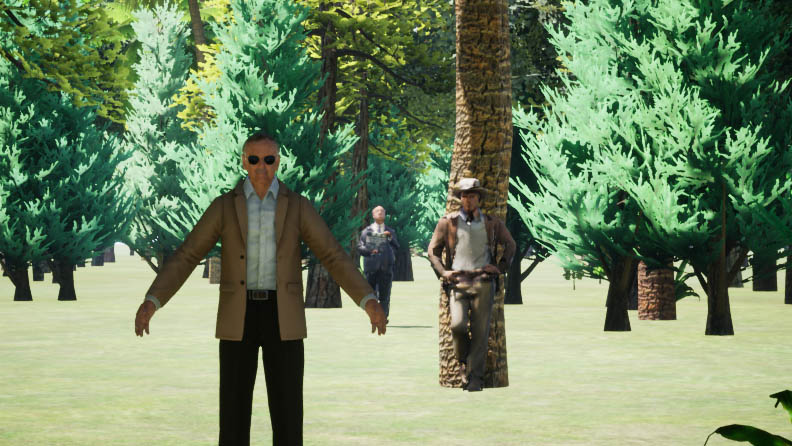}
         &
         \includegraphics[width=0.19\linewidth,height=2.25cm]{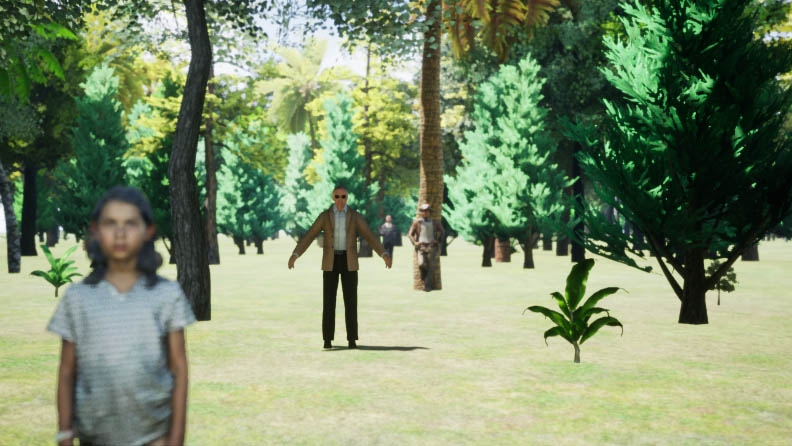}
         &
          \includegraphics[width=0.19\linewidth,height=2.25cm]{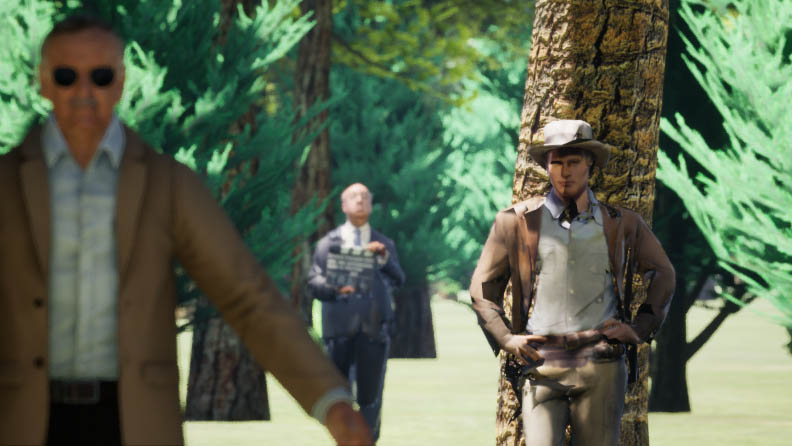}
         &
          \includegraphics[width=0.19\linewidth,height=2.25cm]{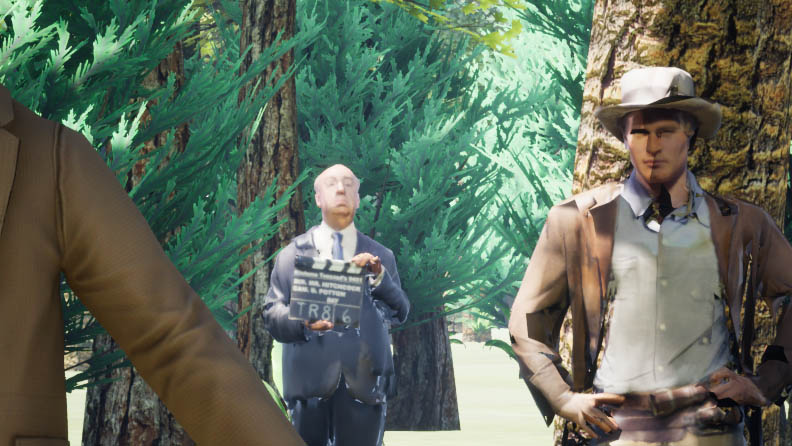}
         \\
         \small Filmback IMAX 70mm &
         \small Lens: 85mm Prime f/1.8 &
         \small Focus distance: 1050cm &
         \small Aperture: f/5.0 &
         \small Focal Length: 300mm\\[5pt]
          \includegraphics[width=0.19\linewidth,height=2.25cm]{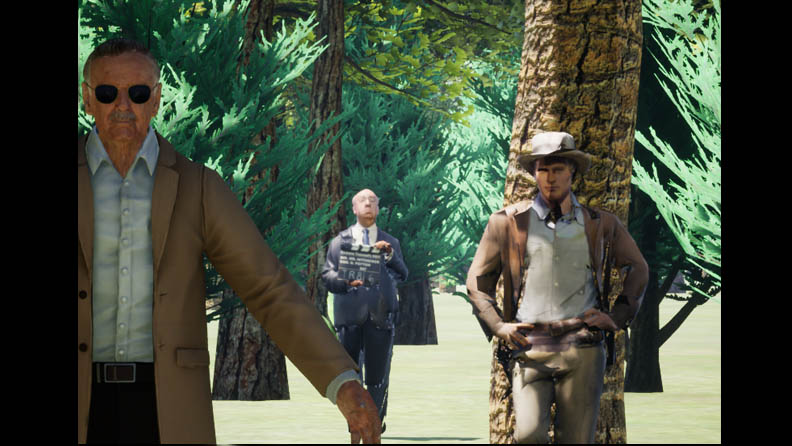}
         &
          \includegraphics[width=0.19\linewidth,height=2.25cm]{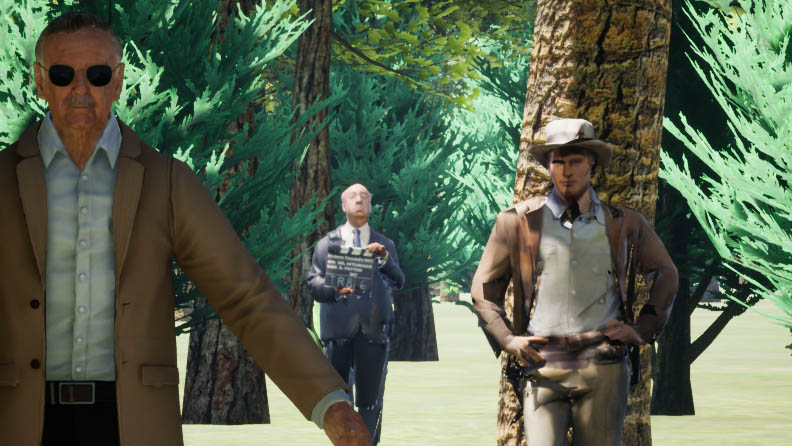}
         &
         \includegraphics[width=0.19\linewidth,height=2.25cm]{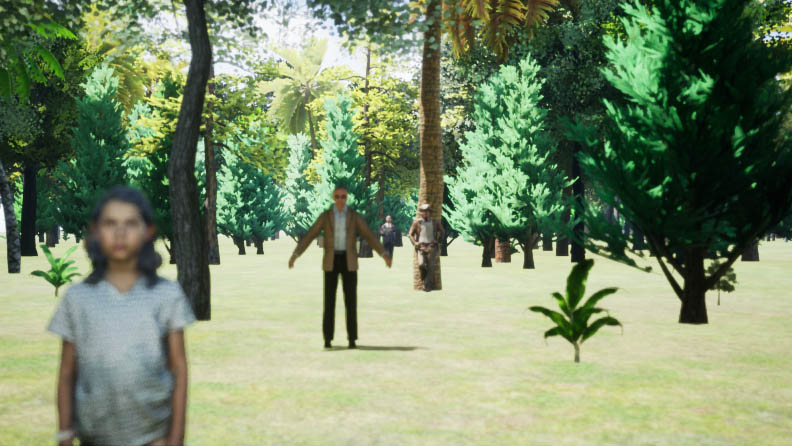}
         &
          \includegraphics[width=0.19\linewidth,height=2.25cm]{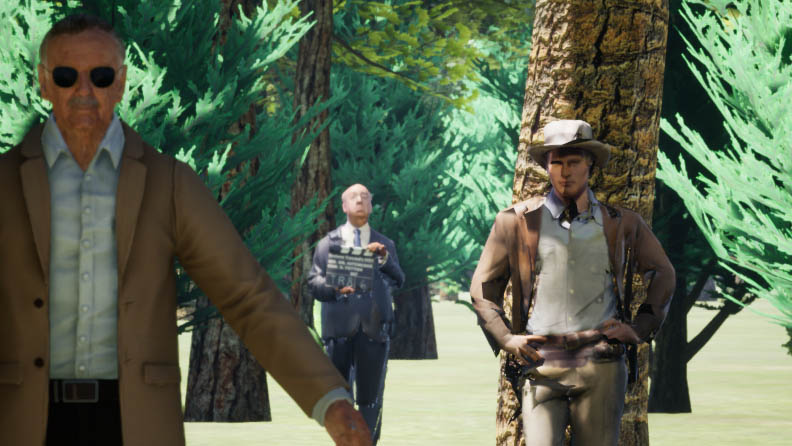}
         &
          \includegraphics[width=0.19\linewidth,height=2.25cm]{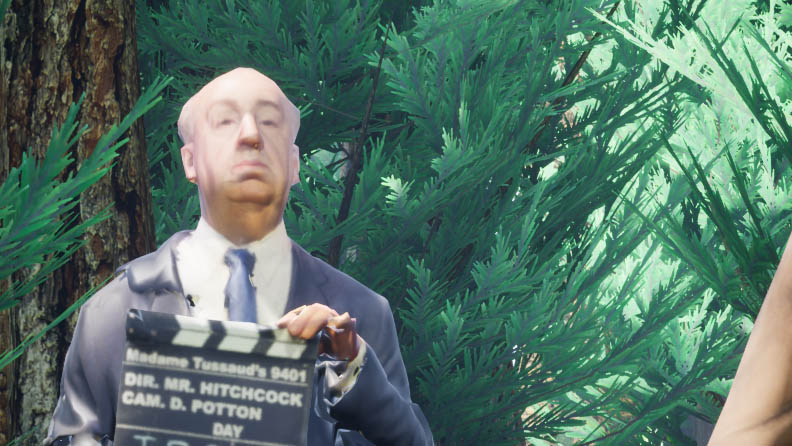}
         \\
         \small Filmback Super 8 mm &
         \small Lens: 200mm Prime f/2 &
         \small Focus distance: 5000cm &
         \small Aperture: f/22.0 &
         \small Focal Length: 800mm\\[5pt]
    \end{tabular}
	\caption{\footnotesize Examples of all the parameters that can be adjusted. First column: Filmback settings. Second column: Lens settings. Third column: Focus distance. Fourth column: Aperture. Fifth column: Focal Length.}
	\label{fig:parameters}
\end{figure*}

In this section we describe the modifications made to AirSim that make the simulation of cinematographic cameras inside the robotics simulator possible. We also describe all of the new functionalities that can be controlled in real-time, with examples to visualize their influence in the resulting images. It is important to note that all the original features of AirSim are available in CinemAirSim, e.g., AirSim's depth and segmentation images.

\subsection{Cinematic Camera Model}
Unreal provides a specialized camera component for cinematic photography referred to as the \emph{CineCameraComponent}~\cite{cinematiccamera_unreal}.
The CineCameraComponent simulates cameras currently used in the film industry in a highly-configurable and realistic way. 
Specifically, it gives users the ability to specify a camera's aperture and focal depth as well as the lens' f-stop and focal length. 
This camera can be modeled by thin-lens camera model whose projection equation is
\begin{equation}
  \frac{1}{f} =  \frac{1}{Z} + \frac{1}{z}
  \, ,
\end{equation}
where \emph{f} represents the focal length, \emph{Z} represents the focus distance (distance of the object to the lens) and \emph{z} represents the focal depth (distance from the lens to the sensor). 
The thin-lens model differs from the common pinhole camera model in that the aperture is not infinitesimally small, which allows multiple rays of light to hit the same pixel on the sensor. This property enables the focal-blur quality of objects that are not at the focus distance for a given focal length and focal depth. 


\subsection{Cinematic Camera Integration}
In order to integrate the cinematic camera model in the architecture of AirSim, we have remodeled some of its modules. The parts modified for CinemAirSim are highlighted in red in Fig.~\ref{fig:software_architecture}. 

Firstly, in the Sensor Simulation module of AirSim, we have replaced the base pinhole cameras, \emph{CameraComponent}, that are embedded in the drones for cinematic cameras, \emph{CineCameraComponent}. Since the weight of the real drone and the mounted camera will be different depending on the commercial models,
we have left this setting with the default values.
Nevertheless, this parameter can be set in the standard configuration files of AirSim.
Secondly, in order to adjust the lens characteristics of the \emph{CineCameraComponent} on-the-fly, we have increased the number of accessors in the class in charge of controlling the camera sensor, \emph{PIPCamera}, implementing all the getters and setters that allow to update these parameters.
The user can still use the methods from the basic cameras of Unreal, as it is just an extended version of the original cameras.

These changes have required other minor adjustments in intermediate files up to the API layer.
The API server and client have been endowed with new methods that give the user the possibility of changing the lens parameters. 
The camera settings can be read and written by making RPC calls as it is done in the base version of AirSim. 
\begin{figure} [!h]
\centering
\includegraphics[width=0.49\linewidth,height=2.8cm]{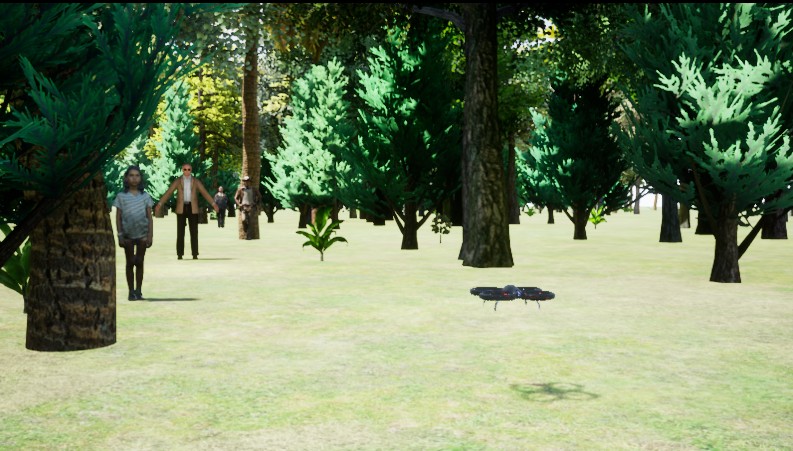}
\includegraphics[width=0.49\linewidth,height=2.8cm]{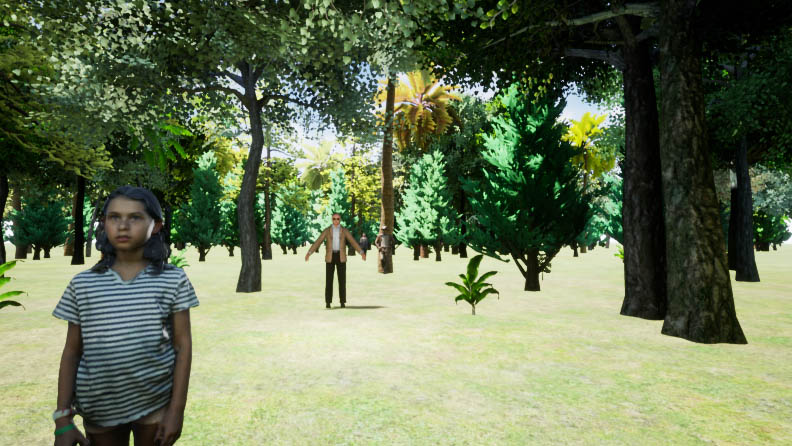}
\caption{\footnotesize Scene with the filming drone (left) and output of the default camera of AirSim (right).}
\label{fig:default_airsim}
\end{figure}

\subsection{Customizable Settings} \label{customizable}
We describe in more detail all the new parameters that can be adjusted using the API.
Their influence in the image is shown with different examples in Fig. ~\ref{fig:parameters}.
They can be compared to Fig. \ref{fig:default_airsim}, that shows the RGB camera output of the base version of AirSim as well as a third person view of the drone.

\subsubsection{Filmback}
This parameter controls the size of the filmback or digital sensor of the camera. CinemAirSim enables the control of this feature either by setting a custom horizontal and vertical size (width and height) in millimetres or to choose among a list of some common commercial filmbacks sizes (16:9 DSLR, 35mm Academy, IMAX 70mm, etc.). Changing the filmback will automatically change the aspect ratio of the images acquired with the camera.

\subsubsection{Commercial lenses}
Each off-the-shelf lens is described by several parameters: a maximum and a minimum of the focal length and aperture and a minimum focus distance, i.e., the distance to the closest object the lens can focus on. Its diaphragm has a determined number of blades, which influences the shape of the Bokeh effect.
The ability to change the lens of the simulator enables the user to test different lenses without the need of buying new cameras.
As in the filmback, there is a predefined set of commercial lenses available, but it is also possible to manually set custom values of the features previously mentioned. 

\subsubsection{Focus distance}
The focus distance is the space in centimeters between the camera and the subject of the plane where the image is in focus. 
The user can decide whether to activate the manual focus or not. If it is activated, it will focus the image in terms of the focus distance, stated by the user through the API. By combining the focus distance and the focus aperture, the scene achieves a more realistic cinematographic feeling. 
Moreover, we have included the option to show the debug focus plane in the visualization. This is a colored plane that is drawn in the image at the distance where the elements are in focus. 
This helps to control where to adjust the focus distance with precision.
Further information regarding Unreal's focus-defocus process can be found in~\cite{depth_of_field_unreal}.

\subsubsection{Focus aperture}
The aperture of the diaphragm controls the amount of light that enters the image sensor and is expressed in terms of the \emph{f-stop} number. The diameter of the opening where the light passes decreases as this number gets higher, allowing less light to reach the sensor.
The quantity of light that reaches the sensor will determine the depth of field, which is the area between the closest and the farthest points of the image that are in focus.
Larger aperture values in terms of f-stop will make the range bigger. 
Choosing different values of focus distance will 
determine the position of the area where the objects will be focused.
By tweaking the focal length and the focus distance, it is possible to get focused outputs from objects that are apparently far.

\subsubsection{Focal length}
The focal length is the distance between the optical center of the lens and the camera's digital sensor, 
expressed in millimeters. In this way it is possible to get closer or farther shots of a desired object without the need of moving the robot. The focal length also controls the field-of-view (FOV) and the zoom in the image. 

\begin{figure} [!h]
\centering
\includegraphics[width=0.49\linewidth,height=2.8cm]{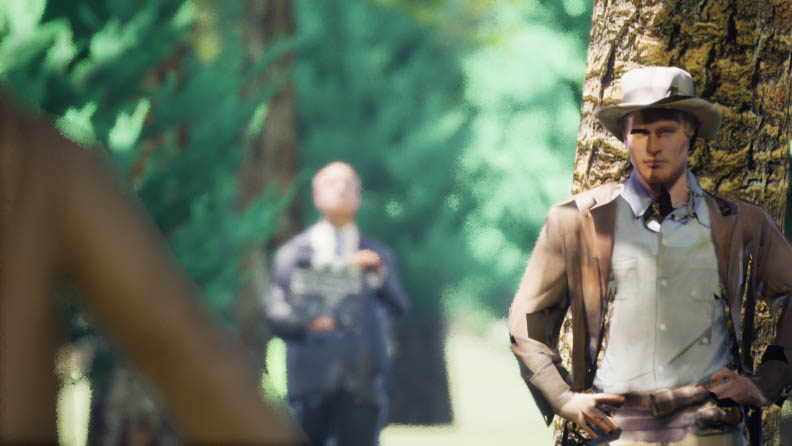}
\includegraphics[width=0.49\linewidth,height=2.8cm]{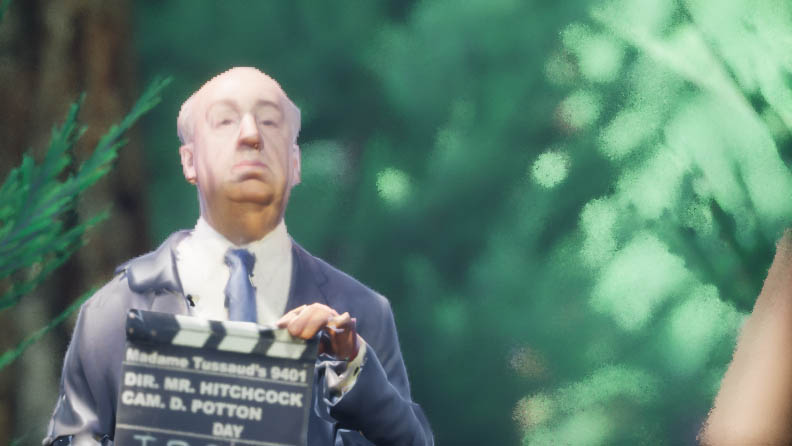}
\caption{\footnotesize Parameters of the camera (left-right): Focus Distance: 2000cm - 3400cm, Focal Length: 300mm - 800mm, Focus Aperture: f/4.0 - f/6.0.}
\label{fig:combined}
\end{figure}

In summary, by combining all the aforementioned parameters, it is possible to achieve a realistic simulation of a scene. Fig. \ref{fig:combined} shows the results achieved by combining changes in the Focal Length, Focus Distance and Aperture of the camera.

\section{Experiments}

\begin{figure} [!h]
\centering
\includegraphics[width=0.49\linewidth,height=2.8cm]{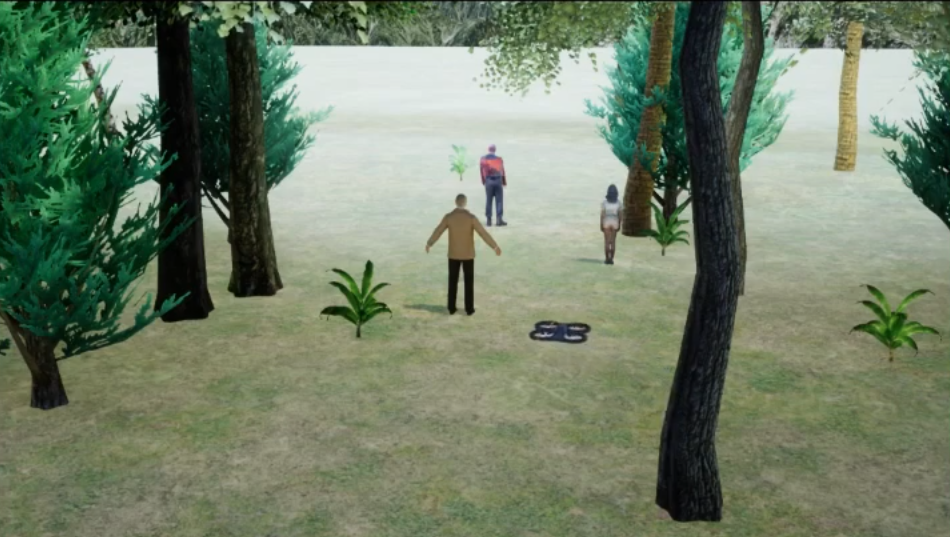}
\includegraphics[width=0.49\linewidth,height=2.8cm]{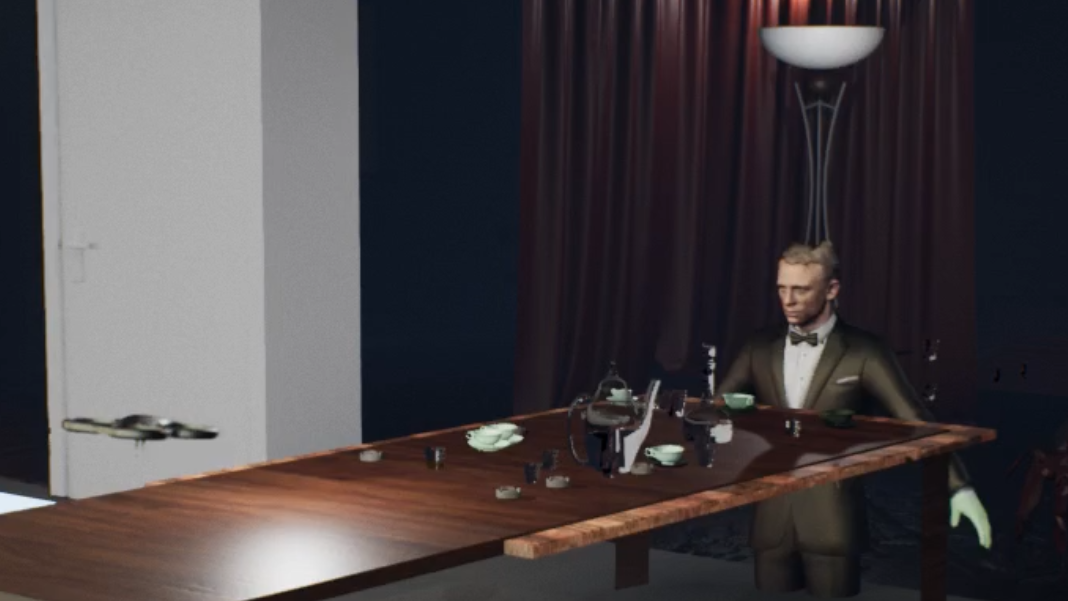}
\caption{\footnotesize Third person view of the scenes with the filming drones.}
\label{fig:thirdPersonDrones}
\end{figure}

In this section, we have reproduced some scenes of films that were awarded with the Academy's Oscar for Best Photography to demonstrate the potential of the simulator and to highlight new avenues of research where it can be used.
We want to remark that the goal of these experiments is not to propose a new planning and control solution for the camera parameters but to demonstrate
that these type of plans can be efficiently tested and validated using the simulator. The experiments provide qualitative results since we are not able to obtain the ground truth configuration of the cameras used to record these scenes.
To obtain these results, in the experiments we have generated open-loop combined trajectories for both the camera and drone parameters that mimic the real scenes filmed with cinematographic cameras. We have compared them with equivalent experiments in which only the position of the drone is updated, using a conventional camera.
Fig.~\ref{fig:thirdPersonDrones} shows the two scenes with the filming drones from a third person view.
We refer the reader to the supplementary video\footnote{\href{https://www.youtube.com/watch?v=A2Ni9dUkJOw}{https://www.youtube.com/watch?v=A2Ni9dUkJOw}}, where all of these scenes, the simulations, and comparisons can be observed in more detail. 

\subsection{Focal length changes}
\label{road2_section}
In the experiment of Fig.~\ref{fig:main}, we have reproduced a scene acquired by a drone where the focal length is controlled.
The scene is extracted from the movie ``The Revenant'' (awarded in 2015).
Changing this parameter affects the field of view, generating a zoom-in or zoom-out effect without the need of getting closer or farther.
There is another experiment of similar characteristics extracted from the movie ``Road To Perdition'' in the supplementary video.

\subsection{Focus distance and aperture changes} 
\label{road1_section}
In this experiment we put the emphasis on the parameters of the cinematographic camera that change the focus of the image, i.e., the aperture and focus distance.
The objective is to provide values for these two parameters in such a way that the objects that are in focus match with the scene from the movie.
The scene we have chosen is extracted from ``Road to Perdition''. 

\begin{figure} [!h]
\centering
	\subfigure{\includegraphics[width=0.32\linewidth,height=1.6cm]{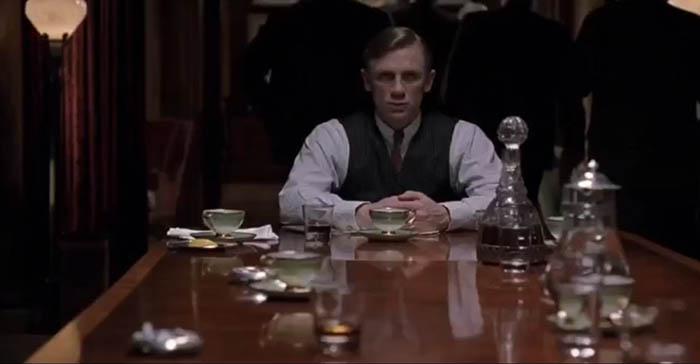}}
	\subfigure{\includegraphics[width=0.32\linewidth,height=1.6cm]{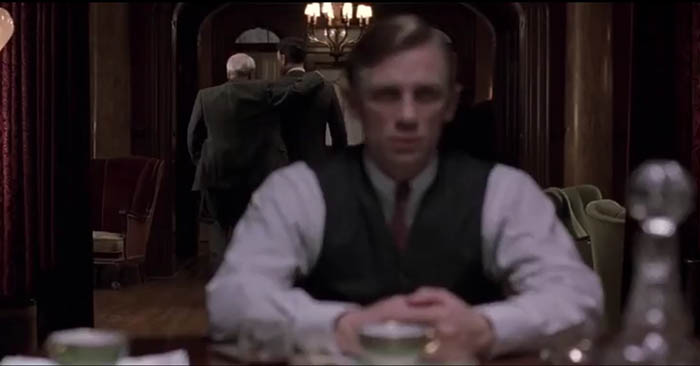}}
	\subfigure{\includegraphics[width=0.32\linewidth,height=1.6cm]{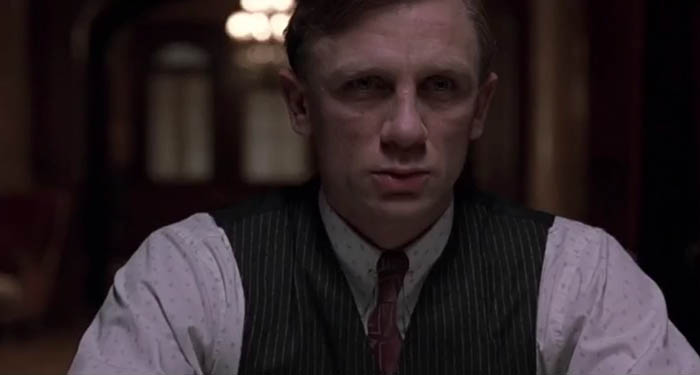}}
	\subfigure{\includegraphics[width=0.32\linewidth,height=1.6cm]{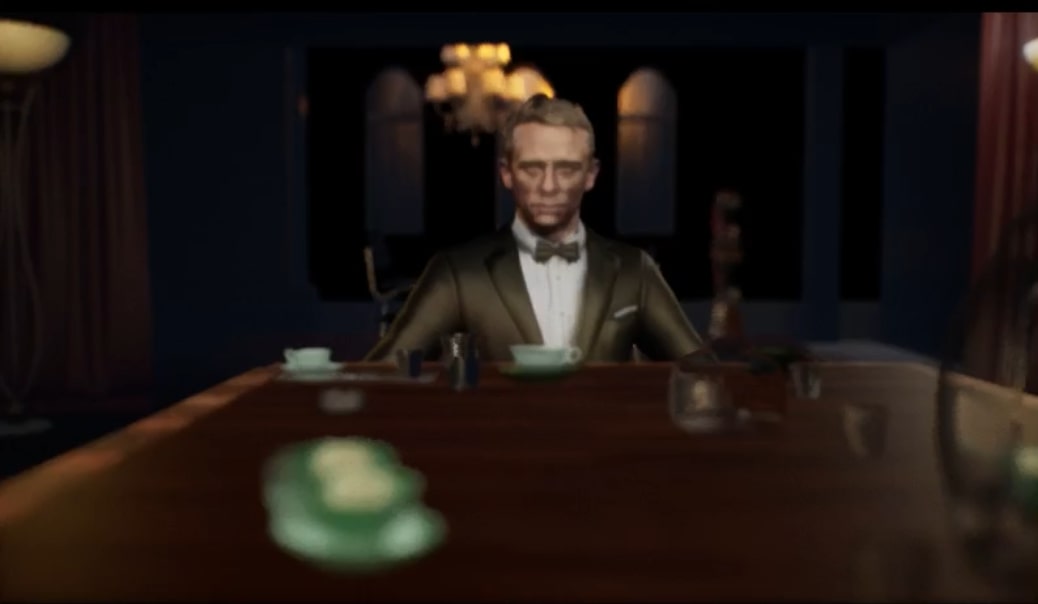}}
	\subfigure{\includegraphics[width=0.32\linewidth,height=1.6cm]{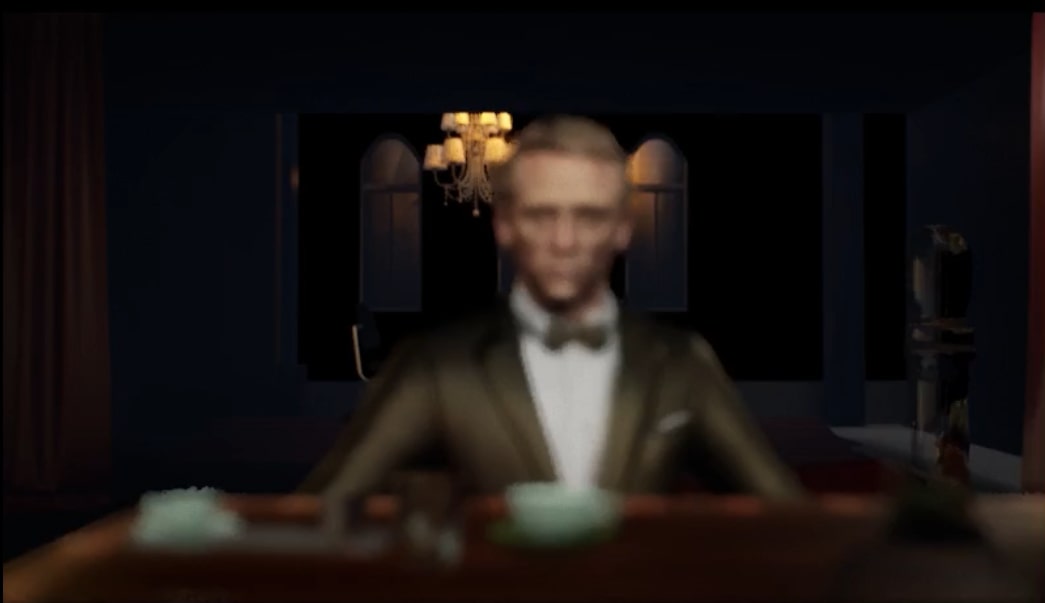}}
	\subfigure{\includegraphics[width=0.32\linewidth,height=1.6cm]{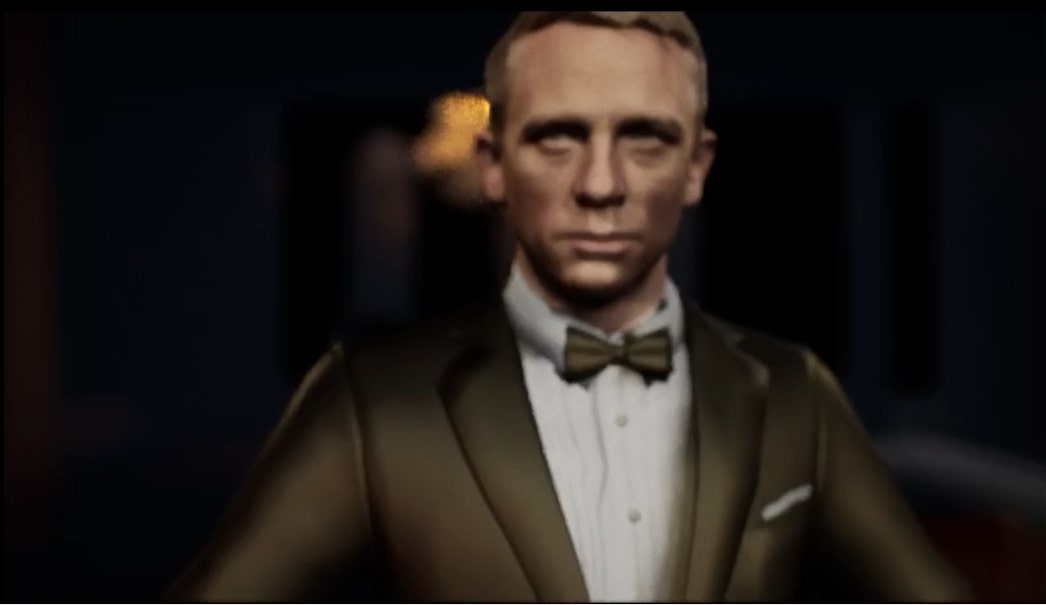}}
	\subfigure{\includegraphics[width=0.32\linewidth,height=1.6cm]{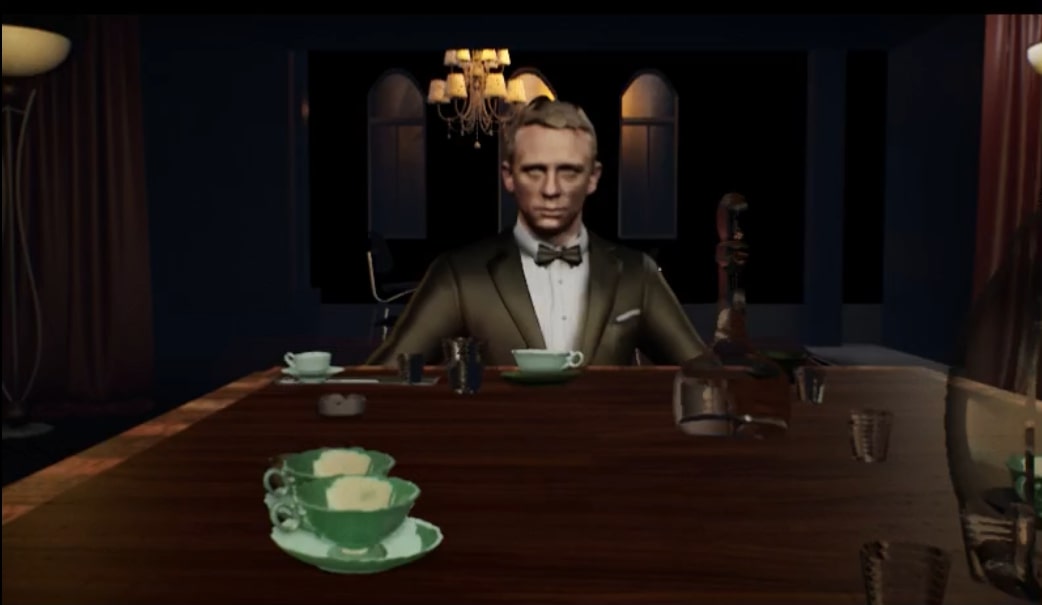}}
	\subfigure{\includegraphics[width=0.32\linewidth,height=1.6cm]{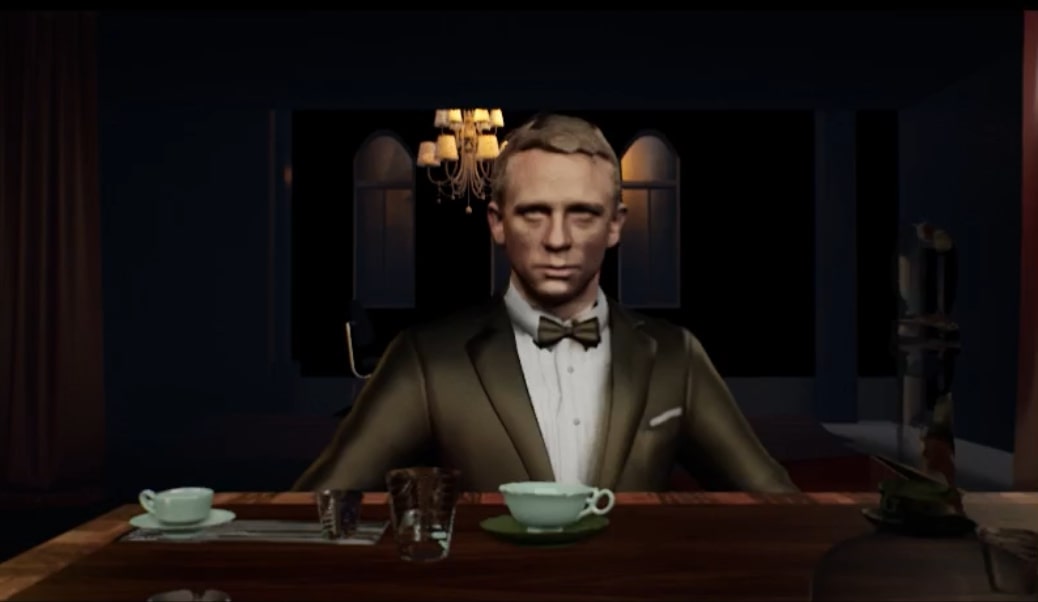}}
	\subfigure{\includegraphics[width=0.32\linewidth,height=1.6cm]{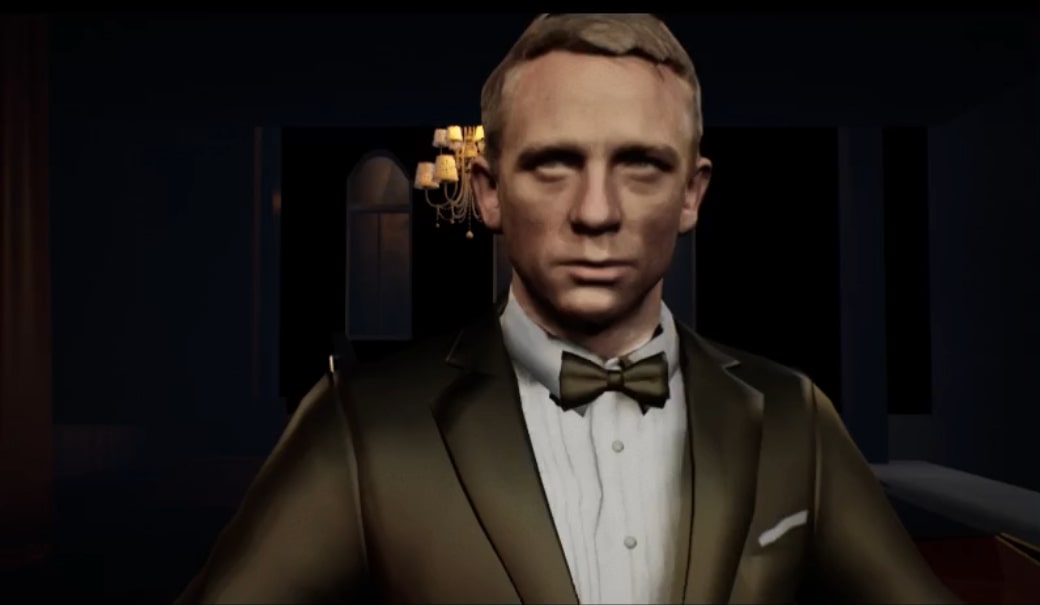}}
\caption{\footnotesize Exemplification of live changes of Focus Distance. Scene from the film Road to Perdition (2002). First row shows frames of the real movie. Second row shows a reproduced version of the scene using CinemAirSim. Third row shows a reproduced version of the scene using AirSim.
}
\label{fig:road_table}
\end{figure}

\begin{figure} [!h]
\centering
	\includegraphics[width=0.7\linewidth]{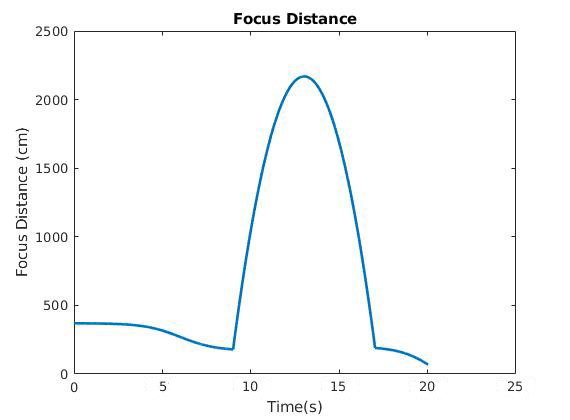}
\caption{\footnotesize Trajectory for the focus distance in the sequence of Fig.~\ref{fig:road_table}.}
\label{fig:road_plots}
\end{figure}

Fig.~\ref{fig:road_table} shows a few select frames from this experiment.
In the left column, the focus is on the main character that is at the end of the table. Close and far away objects, like the green cups and the chandelier respectively, appear out-of-focus. In the bottom row we can see that the standard camera of AirSim cannot reproduce this effect.
As the scene evolves the focus shifts towards the back of the room (middle column) defocusing the main character in the process.
Finally, at the end of the scene (right column), the focus returns to the character.
We have fixed the aperture at a low level, f/1.2. We have used the focus distance trajectory depicted in Fig.~\ref{fig:road_plots}. With the live changes of these parameters, we have achieved a qualitatively similar result between the real and the simulated scene.

\subsection{Focus Distance Closed-loop Control}
\begin{figure} [h]
\centering
	\subfigure{\includegraphics[width=0.50\linewidth,height=3.3cm]{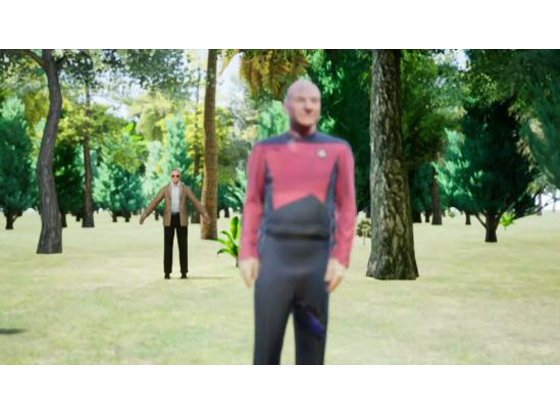}}
	\subfigure{\includegraphics[width=0.48\linewidth,height=3.3cm]{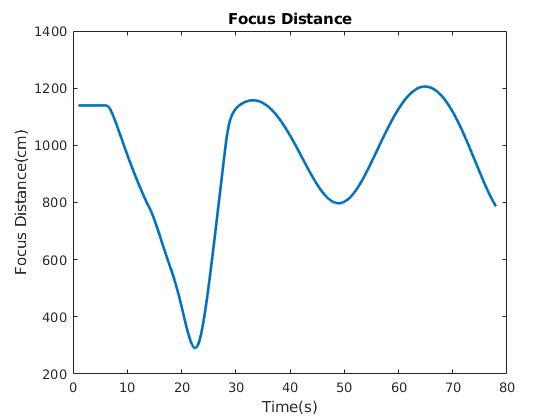}}
	\subfigure{\includegraphics[width=0.50\linewidth,height=3.3cm]{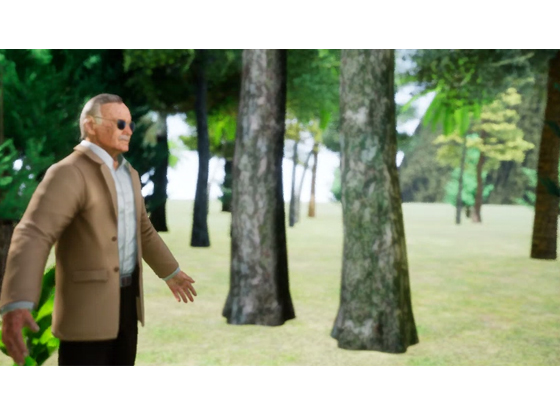}}
	\subfigure{\includegraphics[width=0.48\linewidth,height=3.3cm]{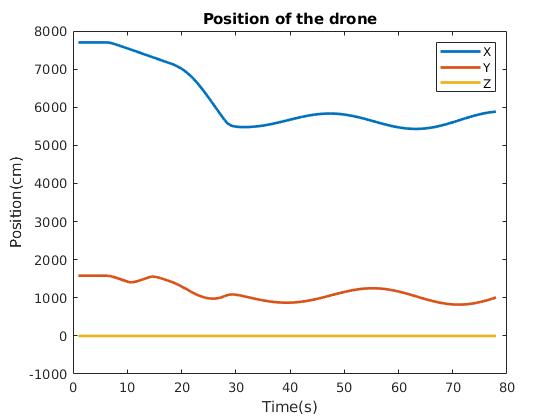}}
	\subfigure{\includegraphics[width=0.50\linewidth,height=3.3cm]{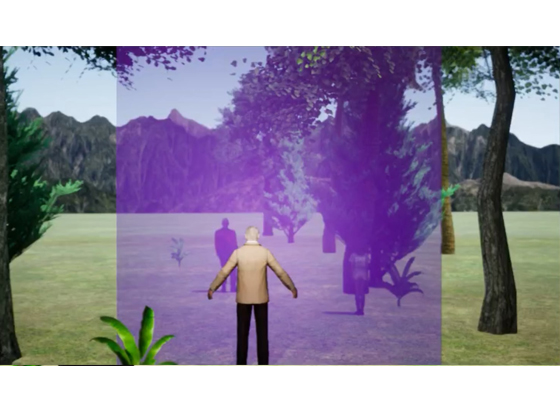}}
	\subfigure{\includegraphics[width=0.48\linewidth,height=3.3cm]{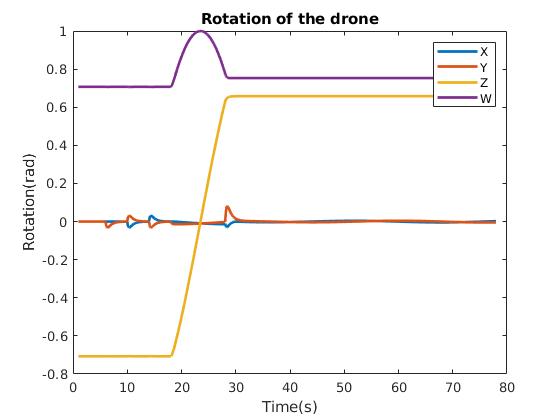}}
\caption{\footnotesize Focus Distance Closed-loop Control. The drone adjusts the focus distance online to keep the target (Stan Lee) in focus during the whole motion. The left column shows the frames acquired by the camera at three instants. The bottom left picture shows the focus plane, that verifies that the target is in focus. The right column shows the focus distance trajectory (top) and the drone position (middle) and orientation (bottom).
}
\label{fig:plots_focus_experiment}
\end{figure}

In this subsection we describe a simple control experiment to adjust the focus distance in terms of the real space that exists between the drone to the object that should be in focus.
For this case, we have chosen the 3D model of Stan Lee to be in focus (Fig.~\ref{fig:plots_focus_experiment}).
The aperture is fixed to a low f-stop value to make the focus change more significative.
We have used the tools from Unreal and AirSim to get the ground-truth location of the drone and the target.
The drone follows a trajectory where its distance to the main character is continuously changing, ranging between far away and close distances. 
During the execution of the trajectory, the focus distance is adjusted in terms of the real distance in meters that separates the drone and the target. This auto focus control occurs in real time.
The left column of Fig.~\ref{fig:plots_focus_experiment} shows some screenshots of the overall result of the experiment.
The plane that represents the part of the image that is in focus is depicted in Fig~\ref{fig:plots_focus_experiment}, bottom left image. The evolution of the focus distance, as well as the position and rotation of the drone are represented in the right column of Fig.~\ref{fig:plots_focus_experiment}.

\section{Conclusions and New Research Possibilities}
In this paper we have presented CinemAirSim, an extended version of AirSim that enables the simulation and control of cinematographic cameras in real time.
We have described the main tools that form the structure of the simulator, namely Unreal and Airsim, and how they have been modified to allow the simulation of a thin-lens camera model.
With CinemAirSim, all the parameters associated to these realistic cameras can be easily controlled in real-time by means of the extended API in the same way as other parameters of the original simulator.
In addition, to illustrate the potential of CinemAirSim,
we have developed a set of experiments where we have provided a single drone open-loop trajectories for pose and camera parameters with the objective of reproducing some scenes of movies recognized for the quality of their photography.
The results demonstrate that the simulator can achieve high levels of reality under different conditions and parameter configurations.

CinemAirSim also paves the way for novel research directions in the fields of robotics and control such as the design of new control algorithms that autonomously decide the best photographic parameters together with the drone trajectories. We can also envision designing new multi-drone coordinated shots, where different robots put the focus on different parts of the scene and exploit other camera parameters to avoid filming each other. 
We believe CinemAirSim will prove to be an important tool in the development of new robotic solutions in the near future.

\bibliographystyle{IEEEtran}
\bibliography{references}
\balance

\end{document}